# A Preliminary Research on Space Situational Awareness Based on Event Cameras

Kun Xiao[1], Pengju Li[2], Guohui Wang[3], Zhi Li[2], Yi Chen[1], Yongfeng Xie[1], Yuqiang Fang[2]
[1] Beijing Institute of Aerospace Systems Engineering, Beijing, China
[2] Space Engineering University, Beijing, China
[3] China Academy of Launch Vehicle Technology, Beijing, China
e-mail: robin_shaun@foxmail.com

*Abstract*—Event camera is a new type of sensor that is different from traditional cameras. Each pixel is triggered asynchronously by an event. The trigger event is the change of the brightness irradiated on the pixel. If the increment or decrement is higher than a certain threshold, the event is output. Compared with traditional cameras, event cameras have the advantages of high temporal resolution, low latency, high dynamic range, low bandwidth and low power consumption. We carried out a series of observation experiments in a simulated space lighting environment. The experimental results show that the event camera can give full play to the above advantages in space situational awareness. This article first introduces the basic principles of the event camera, then analyzes its advantages and disadvantages, then introduces the observation experiment and analyzes the experimental results, and finally, a workflow of space situational awareness based on event cameras is given.

## I. INTRODUCTION

Space situational awareness (SSA) refers to the ability to view, understand and predict the physical location of natural and manmade objects in orbit around the Earth, with the objective of avoiding collisions [1]. At present, SSA is limited by many factors. The optical load based on conventional image sensors has outstanding problems such as large data volume, large system power consumption, low dynamic range, and poor temporal resolution, which cannot meet the imaging needs under harsh light conditions in space. This makes it difficult to realize real-time monitoring of the space situation and take timely and effective countermeasures.

Event cameras are bio-inspired sensors that differ from conventional cameras: Instead of capturing images at a fixed rate, they asynchronously measure per-pixel brightness changes, and output a stream of events that encode the time, location and sign of the brightness changes [2] [3]. Compared with conventional cameras, event cameras have the advantages of high temporal resolution, low latency, high dynamic range, low transmission bandwidth and low power consumption.

Event cameras are used for object tracking [4] [5], surveillance and monitoring [6], and object/gesture recognition [7] [8] [9]. They are also profitable for depth estimation [10] [11], structured light 3D scanning [12], optical flow estimation [13] [14], HDR image reconstruction [15] [16] [17] and Simultaneous Localization and Mapping (SLAM) [18] [19] [20].

On February 23, 2021, the CubeSat equipped with Inivation's event camera was launched by Rocket Lab's Electron rocket. The mission was completed by the University of Western Sydney, the Canberra Space Agency of the University of New South Wales, and the Royal Australian Air Force. It was the first event camera in-orbit test, which is used to verify its continuous star tracking capabilities [21, 22] to explore the possibility of high performance and low power consumption star tracker.

In this paper, we explore the application of event cameras in the field of space situational awareness. This paper will first introduce the principle of event cameras, then analyze its strength and weakness, then introduce the observation experiment carried out in the space illumination simulation environment and analyze the experimental results, and finally present an SSA workflow based on the event camera.

## II. EVENT CAMERA PRINCIPLE AND EVENT REPRESENTATION

Unlike traditional cameras, which acquire the entire image at a specific sampling frequency (such as 30Hz) through an external clock, event cameras output signals asynchronously and independently according to changes of brightness ((log intensity), called "event". The data flow rate of events is continuously changing, and each event represents that the brightness change felt by a certain pixel at a certain moment reaches a threshold, as shown in Fig. 1. This encoding is inspired by the pulsatile nature of biological visual pathways.

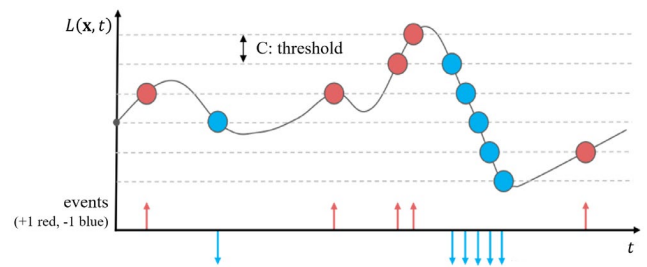

Figure 1. Schematic diagram of event generation from a single pixel

When a pixel emits an event, the current brightness is memorized and changes in brightness are continuously monitored. When the change value exceeds a threshold, the camera generates an event, the event contains the pixel coordinates $x$, $y$, time $t$ and the changed polarity $p$ (+1 for brightness increase, -1 for brightness decrease).

Let the brightness $L(\mathbf{x}, t) = \log I(\mathbf{x}, t)$ be the log intensity of the pixel. When the brightness change of a pixel $\mathbf{x}_k =$

$(x_k, y_k)$ of the event reaches a threshold C, the event camera output an event $\mathbf{e}_k = (\mathbf{x}_k, t_k, p_k)$, as shown in the formula:

$$\Delta L(\mathbf{x}_k, t_k) = L(\mathbf{x}_k, t_k) - L(\mathbf{x}_k, t_k - \Delta t_k) = p_k C$$

where $\Delta t_k$ is the duration from the previous event at the $\mathbf{x}_k$ position to the current event.

Events flow continuously along the time axis, forming the spatiotemporal distribution of events, as shown in Fig. 2. This form is very different from traditional images. In order to adapt to the current SSA architecture, the event stream needs to be processed. The most direct processing method is to superimpose all events in a time window on a two-dimensional image to form an event frame, as shown in Fig. 3. This processing unit is called an event accumulator [19]. However, because events are usually generated from moving edges, the event frame only presents edges of the object, so that a large amount of scene information is lost. Although theoretically, more scene information can be obtained by means of long-time integration to recover the original intensity image, the actual event generation is also related to the upper limit of transmission bandwidth [23] and temperature [24], so the effect of direct integration is limited.

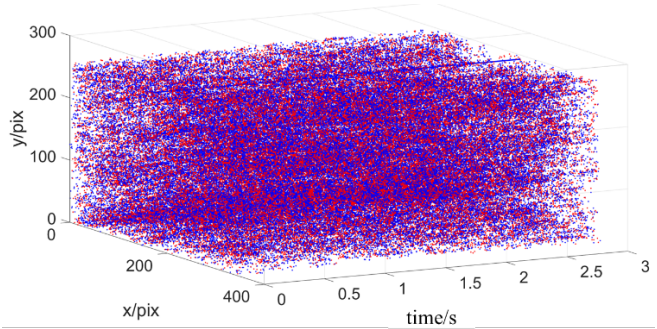

Figure 2. Events in space time, colored according to polarity (positive in blue, negative in red)

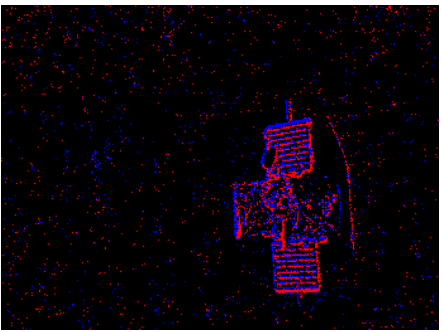

Figure 3. Event frame

The problem of obtaining the intensity image from the event stream is called the image reconstruction problem. The current mainstream solution is deep learning [17], [25], [26]. In this research, E2VID proposed in [17] is chosen in the proposed pipeline for its better performance and good open-source community. The reconstructed intensity image is shown in Fig. 4. Compared to event frames, all texture information is recovered in the intensity image, which meets the need of the current SSA system and supplies input for traditional image processing and computer vision algorithm.

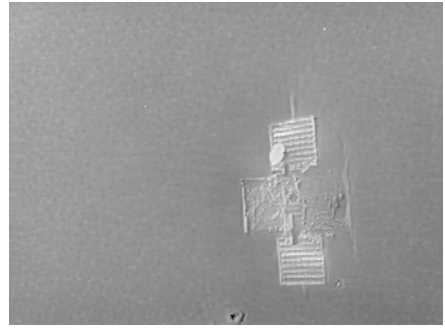

Figure 4. Intensity image reconstructed from events based on deep learning

III. STRENGTHS AND WEAKNESSES OF EVENT CAMERAS

Table 1 summarizes the most popular or recent cameras [27]. The numbers are approximate since they were not measured using a common testbed. Blank cells mean the data are unknown and "NA" means the parameter cannot describe that event camera. The strengths and weaknesses of event cameras will be analyzed below in detail.

*A. Strengths*

**High Temporal Resolution:** Monitoring of brightness changes is fast in analog circuitry, and the read-out of the events is digital, with a 1 MHz clock, i.e., events are detected and timestamped with microsecond resolution. Therefore, event cameras can capture very fast motions, without suffering from motion blur typical of traditional cameras.

**Low Latency**: Each pixel works independently and there is no need to wait for a global exposure time of the frame. As soon as the change is detected, an event is transmitted. Hence, event cameras have minimal latency: about 10 μs on the lab bench, and sub-millisecond in the real world.

**High Dynamic Range (HDR):** The very high dynamic range of event cameras (>120 dB) notably exceeds the 60 dB of high-quality, frame-based cameras, making them able to acquire information from moonlight to daylight. It is due to the facts that the photoreceptors of the pixels operate in logarithmic scale and each pixel works independently, not waiting for a global shutter. Like biological retinas, DVS pixels can adapt to very dark as well as very bright stimuli.

**Low Bandwidth:** Event cameras transmit only brightness changes and thus remove redundant data. For SSA, most pixels reflect the universe background most of the time, so the average bandwidth is much lower than the traditional camera.

**Low Power:** Power is only used to process changing pixels in event cameras. At the die level, most cameras use about 10 mW, and some prototypes achieve less than 10 μW. Embedded event-camera systems where the sensor is directly interfaced to a processor have shown system-level power consumption (i.e., sensing plus processing) of 100 mW or less [28] [29] [30].

## B. Weaknesses

In addition to the significant advantages, there are also some disadvantages in current applications for event cameras.

**Large Noise:** Due to the structure of the sensor, event cameras are sensitive to background activity noise caused by transient noise and leakage currents of semiconductor PN junctions. When the light is low or the sensitivity is high, the background activity will increase, and the noise will be more. [31] [32] study denoising for event cameras, and major manufacturers are optimizing their processes to reduce noise.

**Large Pixel Size:** The pixel size of the current event camera is larger than that of the traditional camera. The pixel size of the event camera shown in Table 1 is mostly above 10 μm, while that of the traditional industrial camera is about 2~4 μm. The large pixel size results in a relatively small resolution of the event camera.

**Small Fill Factor:** Fill factor is the ratio of a pixel's light-sensitive area to its total area. The fill factor of event cameras is commonly small, which means that much pixel area is useless.

TABLE I. COMPARISON OF COMMERCIAL OR PROTOTYPE EVENT CAMERAS

| | Supplier | iniVation | | | Prophesee | | | | Samsung | | | Insightness | CelePixel | |
|---|---|---|---|---|---|---|---|---|---|---|---|---|---|---|
| | Camera model | DVS128 | DAVIS240 | DAVIS346 | ATIS | Gen3 CD | Gen3 ATIS | Gen 4 CD | DVS-Gen2 | DVS-Gen3 | DVS-Gen4 | Rino 3 | CeleX-IV | CeleX-V |
| Sensor specifications | Year | 2008 | 2014 | 2017 | 2011 | 2017 | 2017 | 2020 | 2017 | 2018 | 2020 | 2018 | 2017 | 2019 |
| | Resolution (pixels) | 128×128 | 240×180 | 346×260 | 304×240 | 640×480 | 480×360 | 1280×720 | 640×480 | 640×480 | 1280×960 | 320×262 | 768×640 | 1280×800 |
| | Latency (μs) | 12μs@1klux | 12μs@1klux | 20 | 3 | 40 - 200 | 40 - 200 | 20 - 150 | 65 - 410 | 50 | 150 | 125μs@1klux | 10 | 8 |
| | Dynamic range (dB) | 120 | 120 | 120 | 143 | >120 | >120 | >124 | 90 | 90 | 100 | >100 | 90 | 120 |
| | Min. contrast sensitivity(%) | 17 | 11 | 14.3 - 22.5 | 13 | 12 | 12 | 11 | 9 | 15 | 20 | 15 | 30 | 10 |
| | Power consumption (mW) | 23 | 5 - 14 | 10 - 170 | 50 - 175 | 36 - 95 | 25 - 87 | 32 - 84 | 27 - 50 | 40 | 130 | 20-70 | | 100-450 |
| | Chip size (mm$^2$) | 6.3×6 | 5×5 | 8×6 | 9.9×8.2 | 9.6×7.2 | 9.6×7.2 | 6.22×3.5 | 8×5.8 | 8×5.8 | 8.4×7.6 | 5.3×5.3 | 15.5×15.8 | 14.3×11.6 |
| | Pixel size (μm$^2$) | 40×40 | 18.5×18.5 | 18.5×18.5 | 30×30 | 15×15 | 20×20 | 4.86×4.86 | 9×9 | 9×9 | 4.95×4.95 | 13×13 | 18×18 | 9.8×9.8 |
| | Fill factor (%) | 8.1 | 22 | 22 | 20 | 25 | 20 | >77 | 11 | 12 | 22 | 22 | 8.5 | 8 |
| | Supply voltage (V) | 3.3 | 1.8 & 3.3 | 1.8 & 3.3 | 1.8 & 3.3 | 1.8 | 1.8 | 1.1 & 2.5 | 1.2 & 2.8 | 1.2 & 2.8 | | 1.8 & 3.3 | 1.8 & 3.3 | 1.2 & 2.5 |
| | Stationary noise (ev/pix/s) at 25°C | 0.05 | 0.1 | 0.1 | NA | 0.1 | 0.1 | 0.1 | 0.03 | 0.03 | | 0.1 | 0.15 | 0.2 |
| | CMOS technology (nm) | 350 2P4M | 180 1P6M MIM | 180 1P6M MIM | 180 1P6M | 180 1P6M CIS | 180 1P6M CIS | 90 BI CIS | 90 1P5M BSI | 90 | 65/28 | 180 1P6M CIS | 180 1P6M CIS | 65 CIS |
| | Grayscale output | no | yes | yes | yes | yes | yes | no | no | no | no | yes | yes | yes |
| | Grayscale dynamic range (dB) | NA | 55 | 56.7 | 130 | NA | >100 | NA | NA | NA | NA | 50 | 90 | 120 |
| | Max. frame rate (fps) | NA | 35 | 40 | NA | NA | NA | NA | NA | NA | NA | 30 | 50 | 100 |
| Camera | Max. Bandwidth (Meps) | 1 | 12 | 12 | | 66 | 66 | 1066 | 300 | 600 | 1200 | 20 | 20 | 140 |
| | Interface | USB 2 | USB 2 | USB 3 | | USB 3 | USB 3 | USB 3 | USB 2 | USB 3 | USB 3 | USB 2 | | USB 3 |
| | IMU output | no | 1kHz | 1kHz | no | 1kHz | 1kHz | no | no | 1kHz | no | 1 kHz | no | 1 kHz |

## IV. OBSERVATION EXPERIMENT

In order to test the observation effect of the event camera in the space environment, we carried out a simulated observation experiment in the space lighting simulation environment, and verified the advantages of the event camera compared with the traditional camera in bad light conditions.

### A. Experiment Environment and Devices

Our laboratory has experimental equipment such as a satellite model, a sun simulator, a 3-DoF turntable, a slide rail, a workstation and so on. The laboratory is equipped with a control and analysis system, which can realize observation at any angle, so that the observation experiment can realize automation. A scaled satellite model with a size of 1 meter is used as the observation target. The experiment environment is shown as Fig. 5 and Fig. 6, and the satellite model is shown as Fig.7.

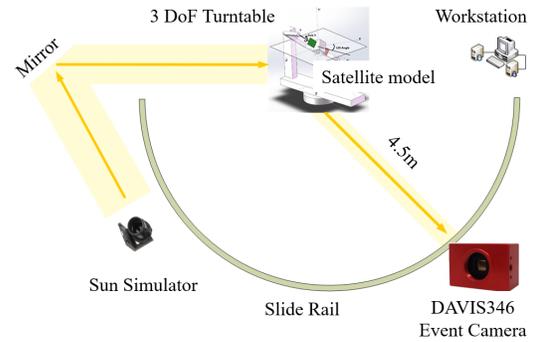

Figure 5. Laboratory equipment

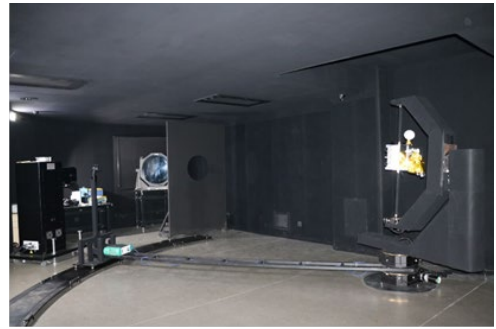

Figure 6. Experiment environment

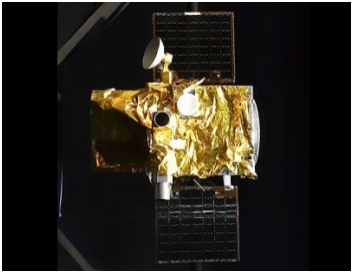

Figure 7. Satellite model

The DAVIS346 event camera from Switzerland's iniVation is selected. The camera has two sensors, which can output events (dynamic vision sensor, DVS) and traditional images (active pixel sensor, APS) simultaneously. DVS and APS share the same optical system, which is convenient for scientific control.

In order to quantitatively measure illuminance, an illuminometer was placed close to the event camera during the experiment. The illuminance of the illuminometer was used to approximately measure that of the camera.

### B. Observation Experiment under Bad Light Conditions

The experiments include high dynamic range and fast motion to test the observation effect of the event camera. As shown in Table 2, the experiments compare traditional images, event frames, and intensity images reconstructed from events based on deep learning.

TABLE II. COMPARISON OF IMAGING BETWEEN TRADITIONAL CAMERAS AND EVENT CAMERAS UNDER BAD LIGHT CONDITIONS

|  | HDR | Extremely HDR | Fast motion |
| --- | --- | --- | --- |
| Illuminance | 5.1 lx | 2370 lx | 0.9 lx |
| Traditional image |  |  |  |
| Event frame |  |  |  |
| Intensity image reconstructed from events |  |  |  |

The images in the first column of Table 2 describe a general HDR scene. Traditional images are partially overexposed, making it impossible to see details in parts with high specular reflectivity, such as solar panels. The event frame and the reconstructed image can clearly show the details of each part. Compared with the event frame, the reconstructed image not only shows the edge information, but also recovers all texture information.

The second column of images depicts a scene where the camera is looking directly at the sun, where the dynamic range of the scene is so high that traditional images can only show sunlight. The event camera captures changes in brightness caused by the movement of the satellite model under sunlight, so the event frame and reconstructed image show the satellite (marked in the yellow box).

The third column of images depicts a scene where the camera moves rapidly in dim light conditions, where the APS requires a long exposure time to image the target, resulting in severe motion blur. The reconstruction algorithm makes full use of the event stream with high temporal resolution and obtains a clear intensity image with sharp edges of the target.

Regarding bandwidth, during the experiment, when the camera does not move relative to the target, the bandwidth of the DVS is only 20kb/s, which comes from noise events. At the same time, the APS still takes photos at a regular rate, resulting in 760kb/s bandwidth. When the camera moves relative to the target, due to the extremely high temporal resolution of the event camera, the output event bandwidth can reach 3Mb/s.

Regarding power consumption, the average power consumption of the DVS is 0.02W, while that of the APS is 0.14W.

### C. Experiment Conclusion

The event camera has the advantage of imaging in HDR scenes, and can even observe the target directly under the sun. In addition, it can image high-speed moving objects without motion blur. Since the event camera only reproduces moving objects, it saves a lot of bandwidth. In addition, the event camera consumes very little power. Because bandwidth and energy are limited in spacecraft, and space light conditions are often poor, the advantages of event cameras can be fully utilized.

## V. SSA WORKFLOW BASED ON EVENT CAMERAS

Since the principles and characteristics of event cameras are quite different from traditional cameras, we design an SSA workflow based on event cameras, as shown in Fig. 8.

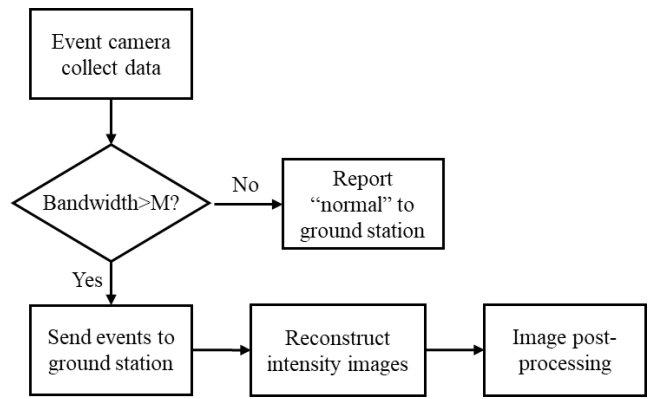

Figure 8. SSA workflow based on event cameras

In order to achieve continuous SSA, the event camera should always be turned on. It consumes a limited amount of power due to its low power consumption. Since there are usually no moving objects in the field of view of the event camera, the bandwidth of the event stream is very low. By setting a threshold M, it can be judged whether the event stream is normal. When the bandwidth is lower than M, it is regarded as a normal state, and the spacecraft only sends the normal state to the ground station without sending event streams, saving space-to-ground communication bandwidth. When the bandwidth is greater than M, it is regarded as an abnormal state. At this time, the spacecraft sends the event stream to the ground station in time. The ground station reconstructs events into images, and then performs post-processing according to the needs of the mission.

As described in Section II, once the event flow has been transformed into high-quality images, traditional computer vision algorithm can be used directly. For example, according to our previous research, we can do 3D reconstruction [33]. Fig. 9 shows the 3D reconstruction pipeline based on the event camera, which shows the 3D reconstruction of the air-bearing spacecraft simulator. It can be seen that after the neural network converts events into images, the classical structure from motion and multi-view stereo technology can be used to achieve 3D dense reconstruction.

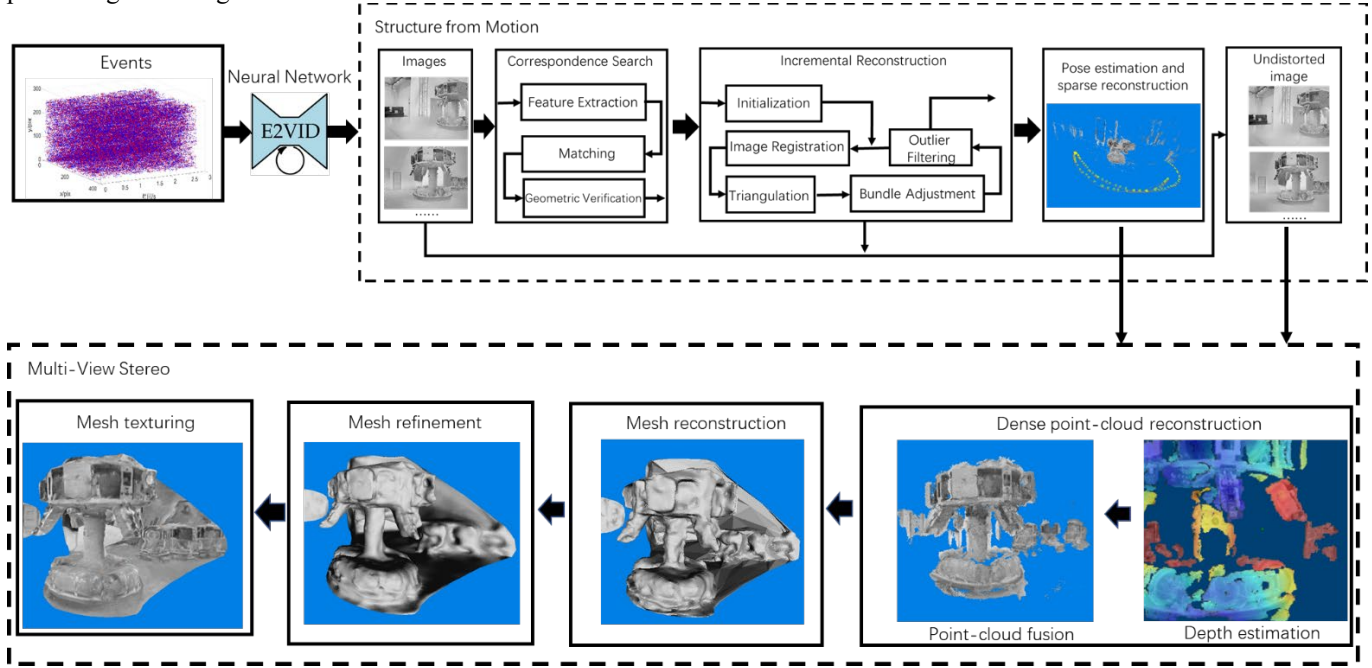

Figure 9. Event-Based Dense Reconstruction Pipeline

Because reconstructing intensity images based on deep learning requires large computing resources but onboard resources are limited, it is carried out on the ground station according to the proposed SSA workflow. However, with the continuous improvement of onboard computing power, it will be possible for spacecraft to reconstruct images from events autonomously and post-process the reconstructed images according to mission requirements in the future.

## VI. Conclusion

This paper analyzes the application of a new type of sensor-event camera for space situational awareness, and designs the workflow. Event cameras have the advantages of high temporal resolution, low latency, high dynamic range, low bandwidth, and low power consumption, and their research and application are developing rapidly. We present this paper hoping that more researchers and engineers will explore the application of event cameras in the astronautic field in the future.